# Adaptive Approach Phase Guidance for a Hypersonic Glider via Reinforcement Meta Learning

Brian Gaudet*
*University of Arizona, 1127 E. Roger Way, Tucson Arizona, 85721*

Kristofer Drozd† and Ryan Meltzer‡
*University of Arizona, 1127 E. Roger Way, Tucson Arizona, 85721*

Roberto Furfaro§
*University of Arizona, 1127 E. Roger Way, Tucson Arizona, 85721*

**We use Reinforcement Meta Learning to optimize an adaptive guidance system suitable for the approach phase of a gliding hypersonic vehicle. Adaptability is achieved by optimizing over a range of off-nominal flight conditions including perturbation of aerodynamic coefficient parameters, actuator failure scenarios, and sensor noise. The system maps observations directly to commanded bank angle and angle of attack rates. These observations include a velocity field tracking error formulated using parallel navigation, but adapted to work over long trajectories where the Earth's curvature must be taken into account. Minimizing the tracking error keeps the curved space line of sight to the target location aligned with the vehicle's velocity vector. The optimized guidance system will then induce trajectories that bring the vehicle to the target location with a high degree of accuracy at the designated terminal speed, while satisfying heating rate, load, and dynamic pressure constraints. We demonstrate the adaptability of the guidance system by testing over flight conditions that were not experienced during optimization. The guidance system's performance is then compared to that of a linear quadratic regulator tracking an optimal trajectory.**

## I. Introduction

Due to the extreme aero-thermal environment associated with hypersonic flight, vehicle structures undergo stress caused by coupling of Fluid-Thermal-Structural Interactions (FTSI) and significant aerodynamic heating typical in hypersonic regimes [1]. This results in non-deterministic time-varying distortions to the airframe and control surfaces that are functions of both peak and integrated heating, which leads to vehicle aerodynamics differing substantially from that assumed during the design of the guidance system. Moreover, the hypersonic flow field exhibits viscous interaction, thin shock layers, an entropy layer, and high-temperature flows that make real gas effects important. Consequently, accurate modeling requires computational fluid dynamics simulators that are orders of magnitude too computationally expensive to use directly for optimizing and testing a guidance and control system. Thus, even without the challenge of structural deformation, the model used for optimization and validation will likely differ substantially from the actual deployment environment. Finally, there is the challenge of a large flight envelope, requiring control over a wide range of dynamic pressure. Clearly, mission risk can be reduced through the use of a guidance system that can adapt in real time to the challenging aero-thermal environment inherent in hypersonic flight that differs substantially from the optimization and testing environment.

Prior work with hypersonic glide guidance in three degrees of freedom that demonstrate robustness to aerodynamic parameter perturbation includes [2], which also respects no fly zones, [3], and [4]. Other published work on 3-DOF glide guidance that does not demonstrate robustness to aerodynamic parameter perturbation includes [2, 5–7]. Most of these implementations required direct control over bank angle and angle of attack, whereas a more realistic system would only give control over bank angle and angle of attack rates, and none of the listed work considers actuator lag, actuator failure, or sensor noise. In [8], the authors consider longitudinal dynamics of an air breathing hypersonic vehicle, and apply reinforcement learning (RL) to disturbance estimation, but do not use RL to optimize the full guidance system. Although there has been recent work in using RL to optimize integrated guidance and control systems for aerospace applications in high fidelity six degrees of freedom (6-DOF) environments [9–12], to date, work in guidance and control for hypersonic vehicles has not taken advantage of recent advancements in RL.

---

*Research Engineer, Department of Systems and Industrial Engineering, E-mail:briangaudet@arizona.edu
†Graduate Student, Department of Systems and Industrial Engineering
‡Graduate Student, Department of Systems and Industrial Engineering
§Professor, Department of Systems and Industrial Engineering, Department of Aerospace and Mechanical Engineering,E-mail:robertof@arizona.edu

In this work we consider a scenario where over a wide range of initial longitude and latitude, the boost-glide hypersonic vehicle separates from a booster at 100 km altitude and an initial speed of 7450 m/s. The guidance system must then guide the vehicle to randomized locations 8000 km from the initial position, at a terminal altitude and speed of 25 km and 1500 m/s respectively. The guidance system is implemented as a policy optimized using reinforcement meta-learning [13–15] (meta-RL) in a three degrees of freedom (3-DOF) simulated environment, leaving the 6-DOF problem for future work. The policy maps observations to commanded bank angle and angle of attack rates, with the observations including a velocity field tracking error formulated using parallel navigation [16], but adapted to work over long trajectories where the Earth's curvature must be taken into account, and modified to allow specification of a trajectory speed profile. Minimizing this tracking error keeps the curved space line of sight to the target location aligned with the vehicle's velocity vector. The optimized guidance system will then induce trajectories that bring the vehicle to the target location with a high degree of accuracy at the designated terminal speed, while satisfying heating rate, load, and dynamic pressure constraints.

In the meta-RL framework, an agent instantiating the policy learns through episodic simulated experience over an ensemble of environments covering the expected distribution of mission scenarios, aerodynamic regimes, sensor and actuator degradation, variation in system time constants, and other factors. The policy is implemented as a deep neural network parameterized by $\theta$ that maps observations to actions $\mathbf{u} = \pi_\theta(\mathbf{o})$, and is optimized using a customized version of proximal policy optimization (PPO) [17]. PPO has been shown to provide state of the art performance in applications ranging from robotic controllers to aerospace integrated guidance and control systems. Adaptation is achieved by including a recurrent network layer [18] with hidden state $\mathbf{h}$ in both the policy and value function networks. Maximizing the PPO objective function requires learning hidden layer parameters $\theta_h$ that result in $\mathbf{h}$ evolving in response to the history of $\mathbf{o}$ and $\mathbf{u}$ in a manner that facilitates fast adaptation to an environment sampled from the ensemble, and generalization to novel environments. The deployed policy will then adapt to off-nominal conditions during flight. Importantly, the network parameters remain fixed during deployment with adaptation occurring through the evolution of $\mathbf{h}$. This is in contrast with the recent work in adaptive control using deep neural networks in an MRAC architecture [19], where parameter learning occurs during deployment, which can be slow, and requires guarantees that a parameter update does not adversely impact stability.

Although in the 3-DOF setting we cannot accurately model hypersonic flight, we attempt to capture the essence of the problem by simulating the vehicle in an environment that at the start of each episode randomly perturbs variables in the dynamics model such as aerodynamic coefficient parameters and atmospheric density. Prior to starting this research, we hypothesized that an integrated guidance and control system based off of a guidance law would be more robust than tracking an optimal trajectory. Our rationale was that an optimal trajectory is only optimal given the dynamics it is optimized with, and can become sub-optimal (and even infeasible) if the deployment environment differs substantially from the optimization environment. In the experiments (Section V) we present evidence that supports this hypothesis, and discuss the issue further in Section V.C. However, it is worth noting that if the optimal trajectory is recomputed at every step of the guidance system, tracking becomes much more robust to off-nominal flight conditions, as was shown in [4]. Nevertheless, it is not clear as to whether this would be feasible real time in six degrees of freedom. Although guidance systems implementing some form of guidance law (typically proportional navigation) are state of the art for homing phase missile guidance, implementing such a guidance law over long trajectories where the curvature of the Earth must be considered can present difficulties. A significant contribution of this work is a method of implementing parallel navigation [16] in curved space in a manner that also allows specifying a speed versus distance profile over the trajectory, and creates a robust and adaptive mapping between navigation outputs and commanded bank angle and angle of attack rates that respects path constraints.

## II. Problem Formulation

### A. Equations of Motion

Figure 1 provides a diagram of the reference frame. The vehicle's position is described in spherical Earth centered coordinates $\mathbf{r} = [R, \theta, \phi]$, where $R$ is the distance between the vehicle and the origin of the Earth centered coordinate system, and $\theta$ and $\phi$ the vehicle's longitude and latitude, respectively. We describe the vehicle's velocity vector $\mathbf{v} = [V, \gamma, \psi]$ in a local vertical local horizontal (LVLH) frame [20] where $V$ is the magnitude of the vehicle's velocity. The flight-path angle $\gamma$ is the angle between the local horizontal plane (i.e., the plane passing through the vehicle and orthogonal to the vector $\mathbf{r}$). The heading angle $\psi$ is the angle between the local parallel of latitude and the projection of $\mathbf{v}$ on the horizontal plane. Combining the position and velocity variables into a state vector, the kinematic (Eq. (1a-1c)) and force equations (Eq. (1d-1f)) for the vehicle are as follows, where $D$ and $L$ are the drag and lift forces, $\sigma$ and $\alpha$ the bank angle and angle of attack. $\Omega = 7.292 \times 10^{-5}$ rad/s is the Earth's angular velocity, and $g = \mu/R$ the Earth's gravitational acceleration, where $\mu$ is Earth's gravitational parameter. $\Delta\sigma$ and $\Delta\alpha$ are the perturbed and lagged commanded change to the bank angle and angle of attack as given in Section II.B.

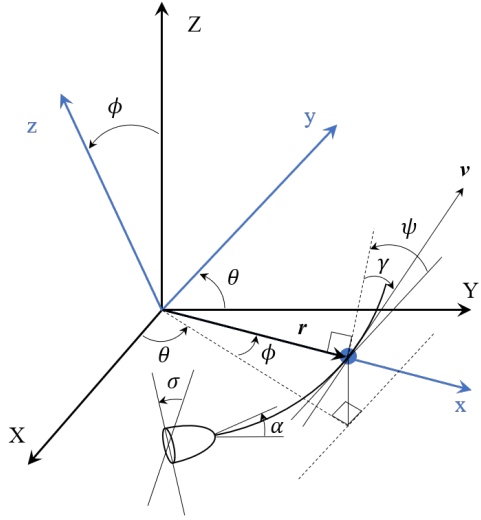

**Fig. 1   Coordinate systems**

$$\frac{dR}{dt} = V \sin(\gamma) \tag{1a}$$

$$\frac{d\theta}{dt} = \frac{V \cos(\gamma) \cos(\psi)}{R \cos(\phi)} \tag{1b}$$

$$\frac{d\phi}{dt} = \frac{V \cos(\gamma) \sin(\psi)}{R} \tag{1c}$$

$$\frac{dV}{dt} = -\frac{D}{m} - g \sin(\gamma) + \Omega^2 R \cos(\phi)(\sin(\gamma)\cos(\phi) - \cos(\gamma)\sin(\phi)\sin(\psi)) \tag{1d}$$

$$V\frac{d\gamma}{dt} = \frac{L \cos(\sigma)}{m} - g \cos(\gamma) + \frac{V^2}{R}\cos(\gamma) + 2\Omega V \cos\phi \cos\psi + \\ \Omega^2 R \cos(\phi)(\cos(\gamma)\cos(\phi) + \sin(\gamma)\sin(\phi)\sin(\psi)) \tag{1e}$$

$$V\frac{d\psi}{dt} = \frac{L \sin(\sigma)}{m \cos(\gamma)} - \frac{V^2}{R}\cos(\gamma)\cos(\psi)\tan(\phi) + 2\Omega V(\tan\gamma \cos\phi \sin\psi - \sin\phi) + \\ \frac{\Omega^2 R \sin(\phi)\cos(\phi)\cos(\psi)}{\cos(\gamma)} \tag{1f}$$

$$\dot{\alpha} = \Delta\alpha \tag{1g}$$

$$\dot{\sigma} = \Delta\sigma \tag{1h}$$

$$D = \frac{\rho V^2 S_{\text{ref}} C_D}{2} \tag{2a}$$

$$L = \frac{\rho V^2 S_{\text{ref}} C_L}{2} \tag{2b}$$

The atmospheric density $\rho$ is calculated using the exponential atmosphere model $\rho = \rho_0 e^{-(r-R_E)/h_s}$, where $\rho_0 = 1.225$ kg/m$^3$ is the density at sea level, $R_E = 6378$ km is the Earth's radius, and $h_o = 7018.00344$ m is the density scale-height. The reference area of the vehicle is $S_{\text{ref}}$, and $C_D$ and $C_L$ are the drag and lift coefficients. These equations of motion are integrated using fourth order Runge-Kutta integration with an integration step size of 1 s. In order to better measure terminal accuracy, once the vehicle is within 10 km of the target, the integration step size is reduced by a factor of 20, which at 1500 m/s gives a minimum achievable average miss distance of 75 m.

### B. Control

The output of the guidance policy $\mathbf{u} = \pi(\mathbf{o}) \in \mathbb{R}^2$ can be split into two components, the commanded changes to the bank angle and angle of attack. These control inputs are then clipped and scaled as shown in Eqs. (3a) through (3b), where $\Delta\alpha_{\max}$ and $\Delta\sigma_{\max}$ are the limits on control input.

$$\Delta\sigma_{\text{cmd}} = \Delta\sigma_{\max}\text{clip}(\mathbf{u}[0], -\Delta\sigma_{\max}, \Delta\sigma_{\max}) \tag{3a}$$

$$\Delta\alpha_{\text{cmd}} = \Delta\alpha_{\max}\text{clip}(\mathbf{u}[1], 0, \Delta\alpha_{\max}) \tag{3b}$$

We model actuator failure and Gaussian actuator noise as shown in Eqs. (4a) and (4b), where in the case of actuator failure, $\epsilon_{\text{ctrl}}$ are the bounds for biasing the control inputs, and $\mathcal{N}(\mu, \sigma, n)$ denotes an n-dimensional Gaussian random variable with mean $\mu$ and standard deviation $\sigma$. $\mathcal{U}(a, b, n)$ denotes an n-dimensional uniformly distributed random variable bounded by $a, b$. Finally, $\sigma_{\text{ctrl}}$ is Gaussian actuator noise.

$$\Delta\sigma_{\text{AF}} = \begin{cases} \Delta\sigma_{\text{cmd}}(1 + \mathcal{U}(\epsilon_{\text{ctrl}}[0], \epsilon_{\text{ctrl}}[1], 1)) + \mathcal{N}(0, \sigma_{\text{ctrl}}, 1), & \text{if } \mathcal{U}(0, 1, 1) < p_{\text{fail}} \\ \Delta\sigma_{\text{cmd}} + \mathcal{N}(0, \sigma_{\text{ctrl}}, 1), & \text{otherwise} \end{cases} \tag{4a}$$

$$\Delta\alpha_{\text{AF}} = \begin{cases} \Delta\alpha_{\text{cmd}}(1 + \mathcal{U}(\epsilon_{\text{ctrl}}[0], \epsilon_{\text{ctrl}}[1], 1)) + \mathcal{N}(0, \sigma_{\text{ctrl}}, 1), & \text{if } \mathcal{U}(0, 1, 1) < p_{\text{fail}} \\ \Delta\alpha_{\text{cmd}} + \mathcal{N}(0, \sigma_{\text{ctrl}}, 1), & \text{otherwise} \end{cases} \tag{4b}$$

Actuator delay is then modeled by integrating Eqs. (5a) through (5b), where $\tau_{\text{ctrl}}$, is the actuator time constant, and $\Delta\sigma$ and $\Delta\alpha$ are the control inputs used in Eqs. (1g) and (1h).

$$\dot{\Delta\sigma} = \frac{\Delta\sigma_{\text{AF}} - \Delta\sigma}{\tau_{\text{ctrl}}} \tag{5a}$$

$$\dot{\Delta\alpha} = \frac{\Delta\alpha_{\text{AF}} - \Delta\alpha}{\tau_{\text{ctrl}}} \tag{5b}$$

### C. Vehicle Model and Mission

As reference model, we chose a configuration similar to the space shuttle [21] and scaled mass and reference area by a factor of 100. We use the drag polar aerodynamic model given in [21], where $C_L$ is a function $\alpha$ (in degrees), $C_D$ a function of $C_L$, and the coefficient values are given in Table 3. Note that angle of attack is held at a constant 40 degrees from the entry interface to $V = 4570$ m/s for thermal protection, after which it varies depending on the commanded change in angle of attack.

$$C_L = C_{L_0} + C_{L_1}\alpha + C_{L_2}\alpha^2 \tag{6}$$

$$C_D = C_{D_0} + C_{D_1}C_L + C_{D_2}C_L^2 \tag{7}$$

We impose path constraints for heating rate $\dot{Q}$, dynamic pressure $q$ and load constraint $n$, with the three path constraints listed in Eqs. (8a) through (8a), where $\dot{Q}_{max} = 3000\text{kW/m}^2$, $q_{max} = 50\text{kN/m}^2$, and $n_{max} = 7g$. Using the Stephan-Boltzmann equation $T_w = \left(\dfrac{\dot{Q}}{\epsilon \sigma_{SB}}\right)^{1/4}$ (where $\epsilon$ is the surface emissivity, and $\sigma_{SB}$ the Stefan-Boltzmann constant), a heating rate of 3000 kW would correspond to a wall temperature of 2808 K. Assuming the vehicle has a carbon aeroshell with a melting point of 3500 K, this leaves significant margin. It is worth noting that any vehicle that would be able to survive the subsequent dive phase would experience a significantly higher heating rate, and would likely need some form of active cooling.

$$\dot{Q} = 1.65 \times 10^{-4} \sqrt{\rho} V^{3.13} \leq \dot{Q}_{max} \tag{8a}$$

$$q = \frac{1}{2}\rho V^2 \leq q_{max} \tag{8b}$$

$$n = \frac{\sqrt{L^2 + D^2}}{m} \leq n_{max} \tag{8c}$$

The vehicle initial state is randomly generated as shown in Table 1, and the goal is to reach the randomly generated target position given in Table 2, with flight time unconstrained. Note that the target longitude is automatically generated so that the initial Haversine distance to target $d$ is as given in Table 1. In Section V we demonstrate that the optimized guidance system can generalize to an extended range of $d_{init}$ by varying initial altitude and flight path angle. The ideal initial heading puts the vehicle on a great circle arc to the target (See Section IV.A), and the heading error perturbs this ideal heading. Note that the ideal heading is actually non ideal when considering the Coriolis force, but works well in practice.

**Table 1    Initial Conditions**

| Parameter | min | max |
|---|---|---|
| Range $d_{init}$ (km) | 8000 | 8000 |
| Altitude $h_{init}$ (km) | 99.8 | 100.2 |
| Speed $V_{init}$ (m/s) | 7430 | 7470 |
| Longitude $\theta_{init}$ (degrees) | 0 | 50 |
| Latitude $\phi_{init}$ (degrees) | -30 | 30 |
| Flight Path Angle $\gamma_{init}$ (degrees) | -0.55 | -0.45 |
| Heading Error $\psi_{error_{init}}$ (degrees) | -0.1 | 0.1 |

**Table 2    Targeted Location**

| Parameter | min | max |
|---|---|---|
| Speed $V_{targ}$ (m/s) | 1500 | 1500 |
| Altitude $h_{targ}$ (km) | 25 | 25 |
| Latitude $\phi_{targ}$ (degrees) | -30 | 30 |

The vehicle parameters are given in Table 3, and Table 4 shows the range over which selected parameters in the dynamics model are perturbed during optimization. Each parameter $\mathbf{p} \in \mathbb{R}^d$ has its value randomly set at the start of an episode as $\mathbf{p} = \mathcal{U}(-v, v, d)$, where $v$ is the bound in the "Value" column of Table 4. This perturbation is meant to capture differences between the dynamics model used for optimization and the actual dynamics encountered in deployment. For example, the aerodynamic coefficients used in optimization will likely not reflect the aerodynamics experienced during deployment due to airframe distortion and surface ablation. The impact of aerodynamic variation is illustrated in Figure 2, where an open loop glide with zero initial heading and flight path error is plotted. The sensor scale factor error is applied to the observation used by the guidance system, specifically $\hat{\mathbf{o}} = \mathbf{o}(1 + \mathcal{U}(-\epsilon_{obs}, \epsilon_{obs}, n))$, where $\mathbf{o} \in \mathcal{R}^n$. We optimize with $\epsilon_{obs} = 0$ in order to test the ability of the optimized policy to generalize to non-zero $\epsilon_{obs}$ in Section V. The guidance system maps observations to actions at a guidance frequency of 0.2 Hz, i.e., the update occurs every 5 seconds.

Table 3  Vehicle Parameters

| Parameter | Value |
|---|---|
| Mass $m$ (kg) | 1043.05 |
| Area Reference $S_{\text{ref}}$ (m$^2$) | 3.9122 |
| Guidance Period (s) | 5 |
| Bank Angle Limits $\sigma_{\max}$ (degrees) | (-150,150) |
| Lift Coefficient Parameters $[C_{L_0}, C_{L_2}, C_{L_3}]$ | [-0.041065 0.016292 0.0002602] |
| Drag Coefficient Parameters $[C_{D_0}, C_{D_2}, C_{D_3}]$ | [0.080505, -0.03026, 0.86495] |
| Angle of Attack Limits $\alpha_{\max}$ (degrees) | (0,40) |
| Angle of Attack Rate Limit $\Delta\alpha_{\max}$ (degrees / s) | 4.0 |
| Bank Angle Rate Limit $\Delta\sigma_{\max}$ (degrees / s) | 10.0 |
| Actuator lag time constant $\tau_{\text{ctrl}}$ (s) | 1 |
| Actuator Noise Standard Deviation $\sigma_{\text{ctrl}}$ (degrees / s) | 0.05 $\Delta\alpha_{\max}$ |
| Actuator Failure Bias Range $\epsilon_{\text{ctrl}}$ | (-0.3, 0.0) |
| Actuator Failure Probability $p_{\text{fail}}$ | 0.5 |
| Sensor Scale Factor Error $\epsilon_{\text{obs}}$ | 0.0 |

Table 4  Dynamics Model Variation

| Parameter | Value |
|---|---|
| Aerodynamic lift coefficient $C_L$ variation | 3% |
| Aerodynamic drag coefficient $C_D$ variation | 3% |
| Atmospheric density $\rho$ variation | 3% |

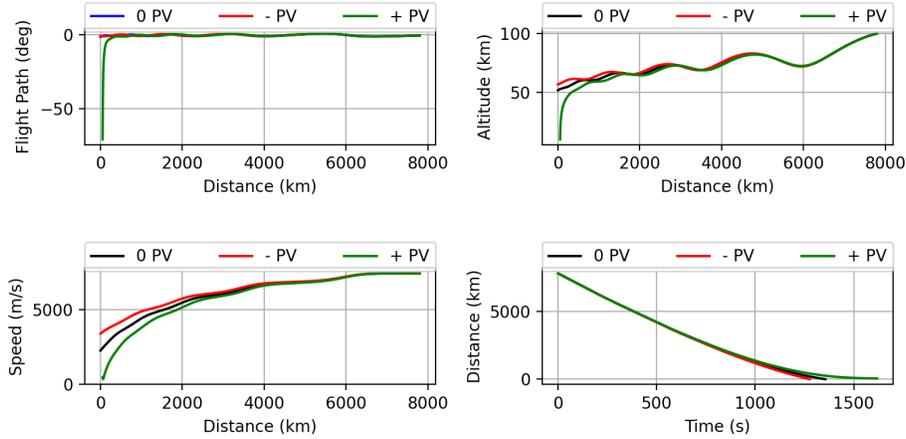

Fig. 2  Open Loop Glide Trajectories Demonstrating Impact of 3% Parameter Variation

## III. Background: Reinforcement Learning Framework

In the reinforcement learning framework, an agent learns through episodic interaction with an environment how to successfully complete a task using a policy that maps observations to actions. The environment initializes an episode by randomly generating a ground truth state, mapping this state to an observation, and passing the observation to the agent. The agent uses this observation to generate an action that is sent to the environment; the environment then uses the action and the current ground truth state to generate the next state and a scalar reward signal. The reward and the observation corresponding to the next state are then passed to the agent. The process repeats until the environment terminates the episode, with the termination signaled to the agent via a done signal. Trajectories collected over a set of episodes (referred to as rollouts) are collected during interaction between the agent and environment, and used to update the policy and value functions. The interface between agent and environment is depicted in Fig. 3.

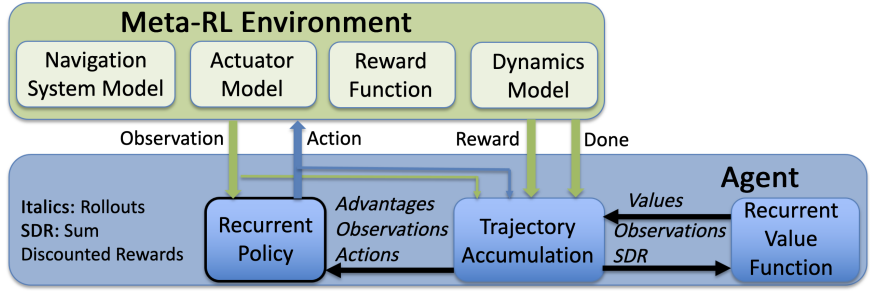

**Fig. 3  Environment-Agent Interface**

A Markov Decision Process (MDP) is an abstraction of the environment, which in a continuous state and action space, can be represented by a state space $\mathcal{S}$, an action space $\mathcal{A}$, a state transition distribution $\mathcal{P}(\mathbf{x}_{t+1}|\mathbf{x}_t, \mathbf{u}_t)$, and a reward function $r = \mathcal{R}(\mathbf{x}_t, \mathbf{u}_t))$, where $\mathbf{x} \in \mathcal{S}$ and $\mathbf{u} \in \mathcal{A}$, and $r$ is a scalar reward signal. We can also define a partially observable MDP (POMDP), where the state $\mathbf{x}$ becomes a hidden state, generating an observation $\mathbf{o}$ using an observation function $O(\mathbf{x})$ that maps states to observations. The POMDP formulation is useful when the observation consists of sensor outputs. In the following, we will refer to both fully observable and partially observable environments as POMDPs, as an MDP can be considered a POMDP with an identity function mapping states to observations.

Meta-RL differs from generic reinforcement learning in that the agent learns over an ensemble of POMPDs. These POMDPs can include different environmental dynamics, aerodynamic coefficients, actuator failure scenarios, mass and inertia tensor variation, and varying amounts of sensor distortion. Optimization within the meta-RL framework results in an agent that can quickly adapt to novel POMDPs, often with just a few steps of interaction with the environment. There are multiple approaches to implementing meta-RL. In [22], the authors design the objective function to explicitly make the model parameters transfer well to new tasks. In [13], the authors demonstrate state of the art performance using temporal convolutions with soft attention. And in [14], the authors use a hierarchy of policies to achieve meta-RL. In this work, we use an approach similar to [15] using a recurrent policy and value function. Note that it is possible to train over a wide range of POMDPs using a non-meta RL algorithm. Although such an approach typically results in a robust policy, the policy cannot adapt in real time to novel environments. In this work, we implement meta-RL using proximal policy optimization (PPO) [17] with both the policy and value function implementing recurrent layers in their networks. After training, although the recurrent policy's network weights are frozen, the hidden state will continue to evolve in response to a sequence of observations and actions, thus making the policy adaptive. In contrast, a policy without a recurrent layer has behavior that is fixed by the network parameters at test time.

The PPO algorithm used in this work is a policy gradient algorithm which has demonstrated state-of-the-art performance for many reinforcement learning benchmark problems. PPO approximates the Trust Region Policy Optimization method [23] by accounting for the policy adjustment constraint with a clipped objective function. The objective function used with PPO can be expressed in terms of the probability ratio $p_k(\boldsymbol{\theta})$ given by,

$$p_k(\boldsymbol{\theta}) = \frac{\pi_{\boldsymbol{\theta}}(\mathbf{u}_k|\mathbf{o}_k)}{\pi_{\boldsymbol{\theta}_{\text{old}}}(\mathbf{u}_k|\mathbf{o}_k)} \tag{9}$$

The PPO objective function is shown in Equations (10a) through (10c). The general idea is to create two surrogate objectives, the first being the probability ratio $p_k(\boldsymbol{\theta})$ multiplied by the advantages $A^{\pi}_{\mathbf{w}}(\mathbf{o}_k, \mathbf{u}_k)$ (see Eq. (11)), and the second a clipped (using clipping parameter $\epsilon$) version of $p_k(\boldsymbol{\theta})$ multiplied by $A^{\pi}_{\mathbf{w}}(\mathbf{o}_k, \mathbf{u}_k)$. The objective to be maximized $J(\boldsymbol{\theta})$ is then the expectation under the trajectories induced by the policy of the lesser of these two surrogate objectives.

$$\text{obj1} = p_k(\boldsymbol{\theta}) A^{\pi}_{\mathbf{w}}(\mathbf{o}_k, \mathbf{u}_k) \tag{10a}$$

$$\text{obj2} = \text{clip}(p_k(\boldsymbol{\theta}) A^{\pi}_{\mathbf{w}}(\mathbf{o}_k, \mathbf{u}_k), 1 - \epsilon, 1 + \epsilon) \tag{10b}$$

$$J(\boldsymbol{\theta}) = \mathbb{E}_{p(\tau)}[\min(\text{obj1}, \text{obj2})] \tag{10c}$$

This clipped objective function has been shown to maintain a bounded Kullback-Leibler (KL) divergence [24] with respect to the policy distributions between updates, which aids convergence by ensuring that the policy does not change drastically between updates. Our implementation of PPO uses an approximation to the advantage function that is the difference between the empirical return and a state value function baseline, as shown in Equation 11, where $\gamma$ is a discount rate and $r$ the reward function, described in Section IV.B.

$$A_{\mathbf{w}}^{\pi}(\mathbf{x}_k, \mathbf{u}_k) = \left[\sum_{\ell=k}^{T} \gamma^{\ell-k} r(\mathbf{o}_\ell, \mathbf{u}_\ell)\right] - V_{\mathbf{w}}^{\pi}(\mathbf{x}_k) \tag{11}$$

Here the value function $V_{\mathbf{w}}^{\pi}$ is learned using the cost function given by

$$L(\mathbf{w}) = \sum_{i=1}^{M} \left( V_{\mathbf{w}}^{\pi}(\mathbf{o}_k^i) - \left[\sum_{\ell=k}^{T} \gamma^{\ell-k} r(\mathbf{u}_\ell^i, \mathbf{o}_\ell^i)\right] \right)^2 \tag{12}$$

In practice, policy gradient algorithms update the policy using a batch of trajectories (roll-outs) collected by interaction with the environment. Each trajectory is associated with a single episode, with a sample from a trajectory collected at step $k$ consisting of observation $\mathbf{o}_k$, action $\mathbf{u}_k$, and reward $r_k(\mathbf{o}_k, \mathbf{u}_k)$. Finally, gradient ascent is performed on $\theta$ and gradient descent on $\mathbf{w}$ and update equations are given by

$$\mathbf{w}^+ = \mathbf{w}^- - \beta_{\mathbf{w}} \nabla_{\mathbf{w}} L(\mathbf{w})|_{\mathbf{w}=\mathbf{w}^-} \tag{13}$$

$$\theta^+ = \theta^- + \beta_\theta \nabla_\theta J(\theta)|_{\theta=\theta^-} \tag{14}$$

where $\beta_{\mathbf{w}}$ and $\beta_\theta$ are the learning rates for the value function, $V_{\mathbf{w}}^{\pi}(\mathbf{o}_k)$, and policy, $\pi_\theta(\mathbf{u}_k|\mathbf{o}_k)$, respectively.

In our implementation of PPO, we adaptively scale the observations and servo both $\epsilon$ and the learning rate to target a KL divergence of 0.001.

## IV. Methods

### A. Guidance Law Formulation

The theoretical basis for our guidance law is taken from parallel navigation [16]. Parallel navigation (constant bearing, decreasing range) attempts to keep the line of sight to target constant for the duration of the engagement. This is equivalent to keeping the vehicle's velocity vector colinear with the line of sight to target [16]. Since in our application we do not have line of sight to target except in the end game, our method calculates new heading and flight path reference angles $\psi_{\text{ref}}$ and $\gamma_{\text{ref}}$ at each step of the guidance law. These reference heading and flight path angles keep the vehicle's velocity vector $\mathbf{v}$ in the LVLH frame aligned with the line of sight to target taking into account the Earth's curvature, using a linear approximation to the curved space line of sight at each step of the guidance law.

Nullifying the heading and flight path errors $\psi_{\text{err}} = \psi - \psi_{\text{ref}}$ and $\gamma_{\text{err}} = \gamma - \gamma_{\text{ref}}$ will create a trajectory that brings the vehicle to the desired target longitude and latitude. However, the mission described in Section II.C requires targeting a terminal speed and altitude $h_{\text{targ}}$. To meet the terminal speed requirement, we construct a reference velocity field $\mathbf{v}_{\text{ref}} = \mathcal{F}(\mathbf{r}_{\text{curr}}, \mathbf{r}_{\text{targ}}) = [V_{\text{ref}}, \psi_{\text{ref}}, \gamma_{\text{ref}}]$ that specifies a full reference velocity rather than a direction vector, with $V_{\text{ref}}$ a speed profile that is a function of Haversine distance to target $d$. $\mathbf{v}_{\text{ref}}$ is used to create shaping rewards, and the terminal reward encourages the agent to meet the terminal altitude requirement, as discussed later in Section IV.B. It might seem that including a reference speed in the guidance law would be a bad idea, as it could induce heating rate and dynamic pressure constraint violations. However, although the meta-RL guidance system is optimized to track this velocity field, it is also optimized to satisfy constraints (See Section IV.B), and can deviate from the reference velocity field to the extent required to satisfy constraints.

In the following, the $\{\cdot\}^U$ superscript denotes the unit sphere reference frame, the $\{\cdot\}^C$ superscript the Cartesian Earth centered reference frame, and no superscript denotes the Earth centered spherical reference frame. The heading reference angle $\psi_{\text{ref}}$ is the heading that keeps the vehicle on a great circle arc from its current position $\mathbf{r}_{\text{curr}}$ to the target position $\mathbf{r}_{\text{targ}}$. To calculate $\psi_{\text{ref}}$, we first transform $\mathbf{r}_{\text{curr}}$ and $\mathbf{r}_{\text{targ}}$ to $\mathbf{r}_{\text{curr}}^U$ and $\mathbf{r}_{\text{targ}}^U$. Next we interpolate a short distance $\text{dr} = 1 \times 10^{-3}$ following the great circle arc along the unit sphere using Eqs. (15a) and (15b), where $\mathbf{r}_{\text{WP}}$ is a waypoint that is a distance dr along the unit sphere great circle arc from $\mathbf{r}_{\text{curr}}^U$ to $\mathbf{r}_{\text{targ}}^U$.

$$\Omega = \arccos\left(\mathbf{r}_{\text{curr}}^U \cdot \mathbf{r}_{\text{targ}}^U\right) \tag{15a}$$

$$\mathbf{r}_{\text{WP}}^U = \mathbf{r}_{\text{curr}}^U \frac{\sin(1-\text{dr})\Omega}{\sin\Omega} + \mathbf{r}_{\text{targ}}^U \frac{\sin(\text{dr})}{\sin(\Omega)} \tag{15b}$$

We then calculate the unit direction vector $\hat{\mathbf{dx}}$ from $\mathbf{r}_{\text{curr}}^{U}$ to $\mathbf{r}_{\text{WP}}^{U}$ and define $\hat{\mathbf{n}} = \mathbf{r}_{\text{curr}}^{U}$ as the normal vector to a plane tangent to the unit sphere at $\mathbf{r}_{\text{curr}}^{U}$ (this plane has the same normal unit vector as the local horizontal plane used to define the LVLH coordinate system). The expression $\hat{\mathbf{dx}}_{\text{PROJ}} = \hat{\mathbf{dx}} - (\hat{\mathbf{dx}} \cdot \hat{\mathbf{n}})\hat{\mathbf{n}}$ is the projection of $\hat{\mathbf{dx}}$ onto the plane with unit normal $\hat{\mathbf{n}}$, and we calculate $\mathbf{v}_{\text{ref}}^{C} = V\hat{\mathbf{dx}}_{\text{PROJ}}$ as the velocity vector that keeps the vehicle on the great circle arc at the vehicle's current speed $V$.

Now let $\hat{\mathbf{z}}$ be the z-axis of the Earth centered reference frame. The intersection of the plane of local latitude and the local horizontal plane can be found as $\hat{\mathbf{x}} = \hat{\mathbf{n}} \times \hat{\mathbf{z}}$. Both the required heading associated with $\mathbf{v}_{\text{ref}}^{C}$ and the heading error $\psi_{\text{err}}$ can then be found as shown in Equations (16a) through (16e).

$$\hat{\mathbf{v}}_{\text{ref}}^{C} = \frac{\mathbf{v}_{\text{ref}}^{C}}{\|\mathbf{v}_{\text{ref}}^{C}\|} \tag{16a}$$

$$|\psi_{\text{ref}}| = \arccos\left(\mathbf{v}_{\text{ref}}^{C} \cdot \hat{\mathbf{x}}\right) \tag{16b}$$

$$\mathbf{c} = \hat{\mathbf{v}}_{\text{ref}}^{C} \times \hat{\mathbf{x}} \tag{16c}$$

$$\psi_{\text{ref}} = \begin{cases} -\psi_{\text{ref}}, & \text{if } \hat{\mathbf{c}} \cdot \hat{\mathbf{x}} > 0 \\ \psi_{\text{ref}}, & \text{otherwise} \end{cases} \tag{16d}$$

$$\psi_{\text{err}} = \psi - \psi_{\text{ref}} \tag{16e}$$

To calculate the reference flight path angle, let $h_{\text{curr}}$ and $h_{\text{targ}}$ be the vehicle's current altitude and target altitude, respectively. Then the reference flight path angle and flight path error are derived via Eqs. (17a) through (17c). It is worth noting that we also experimented with formulating the reference flight path using the quasi-equilibrium guidance condition framework [3], but our approach had superior performance.

$$\text{dh} = h_{\text{curr}} - h_{\text{targ}} \tag{17a}$$

$$\gamma_{\text{ref}} = \arcsin\left(\frac{\text{dh}}{d}\right) \tag{17b}$$

$$\gamma_{\text{err}} = \gamma - \gamma_{\text{ref}} \tag{17c}$$

We can now construct the reference velocity field $\mathbf{v}_{\text{ref}} = \mathcal{F}(\mathbf{r}_{\text{curr}}, \mathbf{r}_{\text{targ}})$ that is a function of the vehicle relative position with respect to the target. Specifically, we first define a reference speed $V_{\text{ref}}$ that is a function of the Haversine distance to target $d$, the vehicle initial speed $V_{\text{init}}$, desired terminal speed $V_{\text{targ}}$, and a distance scaling parameter $\tau$. We then construct a reference velocity vector in Cartesian coordinates $\mathbf{v}_{\text{ref}}^{C}$ and the velocity error $\mathbf{v}_{\text{err}}$ as shown in Equations (18a) through (18b). For an initial distance to target of 8000 km, we found setting $\tau = 2000$ km worked well.

$$V_{\text{ref}} = V_{\text{targ}} + (V_{\text{init}} - V_{\text{targ}})\left(1 - \exp\frac{-d}{\tau}\right) \tag{18a}$$

$$\mathbf{v}_{\text{ref}} = \begin{bmatrix} V_{\text{ref}} & \gamma_{\text{ref}} & \psi_{\text{ref}} \end{bmatrix} \tag{18b}$$

$$\mathbf{v} = \begin{bmatrix} V & \gamma & \psi \end{bmatrix} \tag{18c}$$

$$\mathbf{v}_{\text{err}} = v^{C} - v_{\text{ref}}^{C} \tag{18d}$$

As stated, the guidance law results in trajectories that follow a great circle arc to the target. However, it is also possible to create curved trajectories that result in large lateral diverts. These curved trajectories are calculated by first specifying a series of waypoints between the initial and target positions. Then at each navigation step, cubic splines are used to create a curve starting at the vehicle's current position and passing through the waypoints and target location. The reference heading $\psi_{\text{ref}}$ is then calculated as the angle between the local latitude and the tangent to the curve at the vehicle's current position.

## B. Meta-RL Problem Formulation

An episode terminates if the vehicle altitude falls below 10 km, the heading error is greater than 90 degrees (this occurs during a successful mission as the vehicle flies by the targeted position), or upon violation of the heating rate, dynamic pressure, or load constraints. Trajectories are collected over 60 episodes and used to update the policy and value functions. The agent observation is given in Eq. (19), where $\mathbf{v}_{err}$ is the velocity field tracking error from Eq.(18d), $\psi_{err}$ the heading error from Eq. (16e), $d$ is the Haversine distance between the vehicle's current position and target position, $V$ the vehicle speed, $h_{err}$ the difference between the vehicle current and target altitude, $\alpha$ the angle of attack, and $\sigma$ the bank angle. The action space is $\mathbf{u} \in \mathbb{R}^2$, where each element of $\mathbf{u}$ is scaled and clipped to obtain the commanded rates of change in angle of attack $\Delta\alpha_{cmd}$ and bank angle $\Delta\sigma_{cmd}$, as was discussed in Section II.B.

$$\mathbf{o} = \begin{bmatrix} \mathbf{v}_{err} & d & V & \psi_{err} & h_{err} & \alpha & \sigma \end{bmatrix} \tag{19}$$

The reward function is shown below in Equations (20a) through (20d). $r_{shaping}$ is a shaping reward given at each step in an episode. These shaping rewards take the form of a Gaussian-like function of the norm of the velocity field tracking error $\mathbf{v}_{err}$. $r_{ctrl}$ is a control effort penalty, again given at each step in an episode, and $r_{bonus}$ is a bonus given at the end of an episode if certain conditions are met. Here $d$ is the Haversine distance between the vehicle and target longitude and latitude at the end of the episode (calculated at the target altitude), and $h_{err}$ is the difference between the targeted altitude and vehicle altitude at the end of the episode. Importantly, the current episode is terminated if a constraint is violated, in which case the stream of positive shaping rewards is terminated, and the agent does not receive the terminal reward. It turns out that this is enough to incentivize the agent to meet the constraints, and it is not necessary to give the agent a negative reward when a constraint is violated. We use $\alpha = 1$, $\beta = -0.1$, $\epsilon = 20$, $r_{lim} = 1000$m, $h_{lim} = 1000$m, $\sigma_v = 0.1V$.

$$r_{shaping} = \alpha \exp\left(\frac{-\|\mathbf{v}_{err}\|^2}{\sigma_v^2}\right) \tag{20a}$$

$$r_{ctrl} = \left\| \begin{bmatrix} \frac{\Delta\alpha_{cmd}}{\Delta\alpha_{max}} & \frac{\Delta\sigma_{cmd}}{\Delta\sigma_{max}} \end{bmatrix} \right\| \tag{20b}$$

$$r_{bonus} = \begin{cases} \epsilon, & \text{if } d < r_{lim} \text{ and } |h_{err}| < h_{lim} \text{ and done} \\ 0, & \text{otherwise} \end{cases} \tag{20c}$$

$$r = r_{shaping} + r_{ctrl} + r_{bonus} \tag{20d}$$

The policy and value functions are implemented using four layer neural networks with tanh activations on each hidden layer. Layer 2 for the policy and value function is a recurrent layer implemented using gated recurrent units [18]. The network architectures are as shown in Table 5, where $n_{hi}$ is the number of units in layer $i$, obs_dim is the observation dimension, and act_dim is the action dimension. The policy and value functions are periodically updated during optimization after accumulating trajectory rollouts of 60 simulated episodes.

Table 5  Policy and Value Function network architecture

|  | Policy Network | | Value Network | |
| --- | --- | --- | --- | --- |
| Layer | # units | activation | # units | activation |
| hidden 1 | $10 * \text{obs\_dim}$ | tanh | $10 * \text{obs\_dim}$ | tanh |
| hidden 2 | $\sqrt{n_{h1} * n_{h3}}$ | tanh | $\sqrt{n_{h1} * n_{h3}}$ | tanh |
| hidden 3 | $10 * \text{act\_dim}$ | tanh | 5 | tanh |
| output | act_dim | linear | 1 | linear |

## C. Optimization

Optimization uses the initial conditions and vehicle parameters given in Section II.C. At first the agent learns to avoid violating the heating rate constraint, which leads to longer trajectories, allowing the agent to enter regions of state space where dynamic pressure and load constraints are violated. After around 10,000 episodes of learning, the agent learns to satisfy all constraints, after which the agent adjusts its policy to maximize both shaping and terminal rewards while continuing to satisfy constraints. Learning curves are given in Figures 4 and 5. Fig. 4 plots the mean ("Mean R"), mean minus 1 standard deviation ("SD R"), and minimum ("Min R") rewards on the primary y-axis, with the mean and maximum number of steps per episode plotted on the secondary y-axis. Similarly, Fig. 5 plots terminal miss statistics, with the miss computed as the terminal Haversine distance between vehicle and target measured at the target altitude. These statistics are computed over a batch of rollouts (60 episodes).

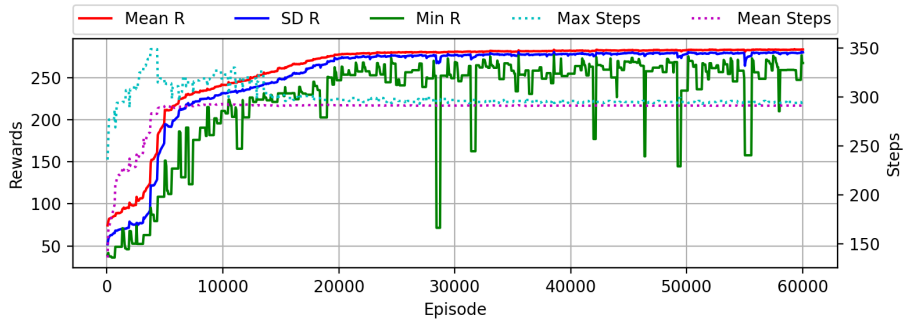

Fig. 4   Optimization Reward History

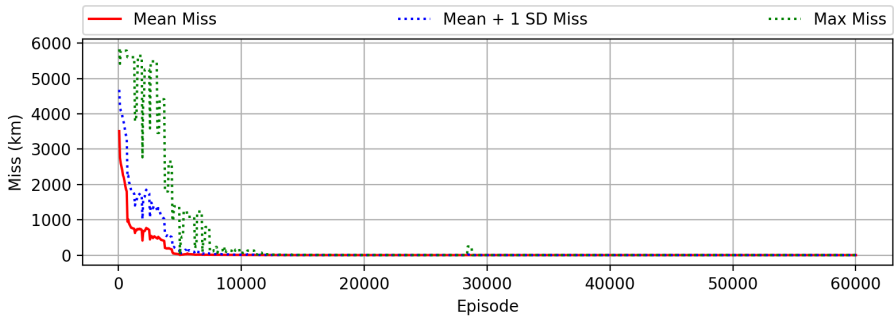

Fig. 5   Optimization Miss History

## V. Experiments

Once the Meta-RL policy was optimized, we conducted multiple experiments to determine the performance of the Meta-RL guidance system and analyze its generalization capability. For the reader's convenience the labels and descriptions of all of the test scenarios we carried out are given in Table 6. Note that in order to allow separation of accuracy metrics from constraint satisfaction, the path constraints were inactive during testing, i.e., a constraint violation did not terminate the episode. However, constraint violations were monitored. The "Optim" case is identical to the conditions used for optimization. The rows in Table 6 following the "Optim" row test the ability of the policy to generalize to conditions not experienced during optimization.

We measure performance using three metrics. The first metric is the terminal Haversine distance between vehicle and target measured at the target altitude, the second metric is the difference between the vehicle terminal altitude and and targeted terminal altitude, and the third metric is the difference between the targeted terminal speed and the actual targeted speed. In the following, these metrics are referred to as the *miss distance*, *altitude error*, and *speed error*, respectively.

### A. Performance of Meta-RL Guidance System

To test the meta-RL guidance system, we ran 5000 episodes over the scenarios given in Table 6. Statistics for terminal miss distance, magnitude of terminal altitude error, and magnitude of terminal speed error are tabulated in

Table 6   Experiment Descriptions

| Label | Description |
| --- | --- |
| Ideal | Ideal conditions are modeled: no parameter variation, actuator failure, sensor noise, actuator noise, or initial state/condition variation. |
| ICV | Similar to the "Ideal" experiment, except that initial state/conditions are varied according to Table 1. |
| Optim | Models parameter variation, actuator failure, actuator noise, and initial state/condition variation Tables 3 and 4, i.e., the same conditions used for optimization. |
| Divert $\Delta$-T | Similar to the "Optim" experiment, but with a random divert maneuver T seconds into the the trajectory, where the target location longitude and latitude are randomly shifted by +/- $\Delta$ degrees, i.e., $\theta_{\text{divert}} = \theta + \mathcal{U}(-\Delta, \Delta, 1)$ and $\phi_{\text{divert}} = \phi + \mathcal{U}(-\Delta, \Delta, 1)$. |
| PV=$\Delta$ | Branches off of the "Optim" experiment, but perturbs aerodynamic parameters and atmospheric density by +/- $\Delta$% rather than 3% |
| Short | Similar to the "Optim" experiment, except that the initial distance to target is set to 6500 km |
| SN=1 | Stems from the "Optim" case, except sensor noise is included where $\epsilon_{\text{obs}} = 0.01$ (see Section II.C). |
| MV/SV=10% | Similar to the "Optim" experiment, but with the vehicle mass and reference area randomly biased in the range +/-10% at the start of each episode. |
| Optim-noAF | Similar to the "Optim" experiment, but no actuator failure. This is only used for the LQR benchmark, as performance deteriorated significantly when actuator failure was considered. |

Table 7. The last two columns are highlighted due to their importance; they indicate the success rate and percentage of constraint violations. We consider a mission successful if the miss distance is less than 1 km and the altitude error is less than 1 km. The rows in Table 7 following the "Optim" row are associated with conditions not experienced during optimization, and we see that performance degrades. The only scenario resulting in constraint violations is the PV=10 case. Note that it would be possible to extend the range of initial distance to target beyond that tested in these experiments by varying initial altitude and flight path angle.

Table 8 gives statistics for the heating rate, load, and dynamic pressure constraints for the "Optim" case. Figure 6 shows a miss distance scatter plot in the "North-East-Down" (NED) frame and and histograms for terminal altitude error and speed for the "Optim" case. Note that the NED frame is centered at the target longitude and latitude, and is therefore unique for each episode. The asymmetric scatter distribution is due to terminating the episode when heading error is greater than 90 degrees, which occurs when the vehicle passes the target location. Figure 7 illustrates a sample trajectory. The transient heading error is due to the fact that keeping the curved space line of sight to target aligned with the reference velocity field requires keeping the vehicle speed consistent with that of the velocity field. Specifically, to reduce speed, the vehicle must bank, which also induces a heading error, which must be corrected, which in turn creates a speed tracking error, and so on. Note that the terminal spikes in heading and flight path error, as well as velocity field tracking error, occur as the vehicle flies past the target location, terminating the episode. Finally, Figure 8 illustrates the evolution of heating, dynamic pressure, and load for the same trajectory shown in Figure 7.

Table 7  Meta-RL Performance

| Case | Miss (m) | | | |Alt Err (m)| | | | $|V_{\text{err}}|$ (m/s) | | | Success | Violation |
| --- | --- | --- | --- | --- | --- | --- | --- | --- | --- | --- | --- |
| - | $\mu$ | $\sigma$ | Max | $\mu$ | $\sigma$ | Max | $\mu$ | $\sigma$ | Max | % | % |
| Ideal | 19 | 0 | 19 | 26 | 0 | 26 | 18 | 0 | 18 | 100.0 | 0.0 |
| ICV | 47 | 20 | 99 | 13 | 43 | 191 | 33 | 17 | 110 | 100.0 | 0.0 |
| Optim | 47 | 20 | 135 | 40 | 95 | 585 | 44 | 37 | 227 | 100.0 | 0.0 |
| Divert=1.0-480 | 49 | 23 | 369 | 20 | 117 | 2369 | 40 | 44 | 279 | 99.88 | 0.0 |
| Divert=1.0-720 | 49 | 23 | 470 | 48 | 107 | 797 | 49 | 43 | 260 | 100.0 | 0.0 |
| PV=5 | 50 | 26 | 1047 | 47 | 127 | 747 | 46 | 51 | 424 | 99.98 | 0.0 |
| PV=10 | 74 | 100 | 3891 | 70 | 263 | 6658 | 352 | 96 | 1274 | 99.90 | 0.1 |
| Short | 49 | 21 | 151 | 10 | 105 | 763 | 33 | 42 | 206 | 100.0 | 0.0 |
| SN=1 | 48 | 26 | 1178 | 34 | 99 | 589 | 42 | 38 | 229 | 99.98 | 0.0 |
| MV/SV=10% | 53 | 23 | 164 | 42 | 100 | 556 | 44 | 57 | 276 | 100.0 | 0.0 |

Table 8  Meta-RL Guidance Constraint Statistics ("Optim" Case)

| Constraint | $\mu$ | $\sigma$ | Max | Limit |
| --- | --- | --- | --- | --- |
| Heating Rate (KW/s) | 678 | 258 | 2575 | 3000 |
| Load (m/s$^2$) | 6.5 | 5.1 | 47.0 | 68.7 |
| Dynamic Pressure (Pa) | 4582 | 8042 | 42635 | 50000 |

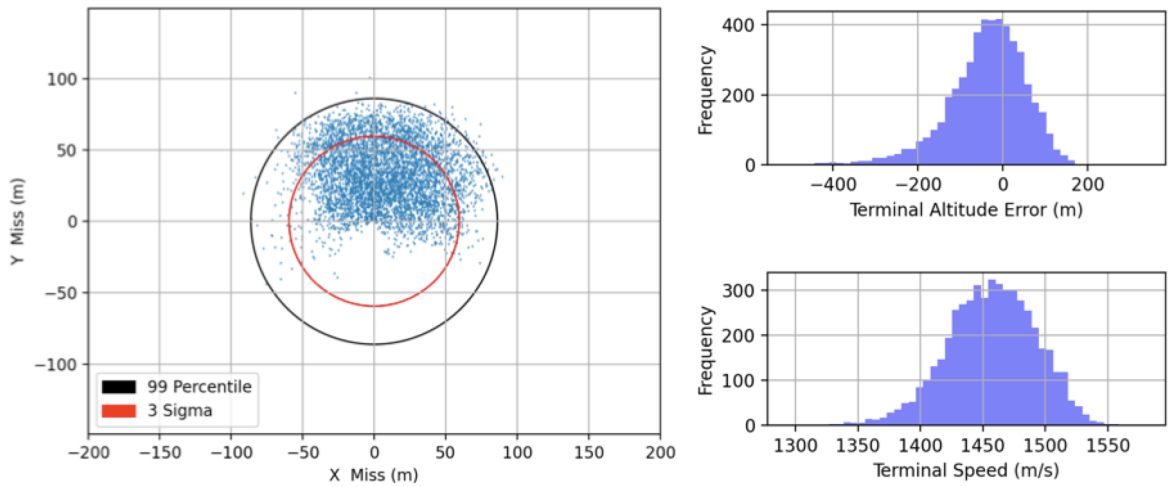

Fig. 6  Meta-RL Guidance Accuracy ("Optim" Case, NED frame)

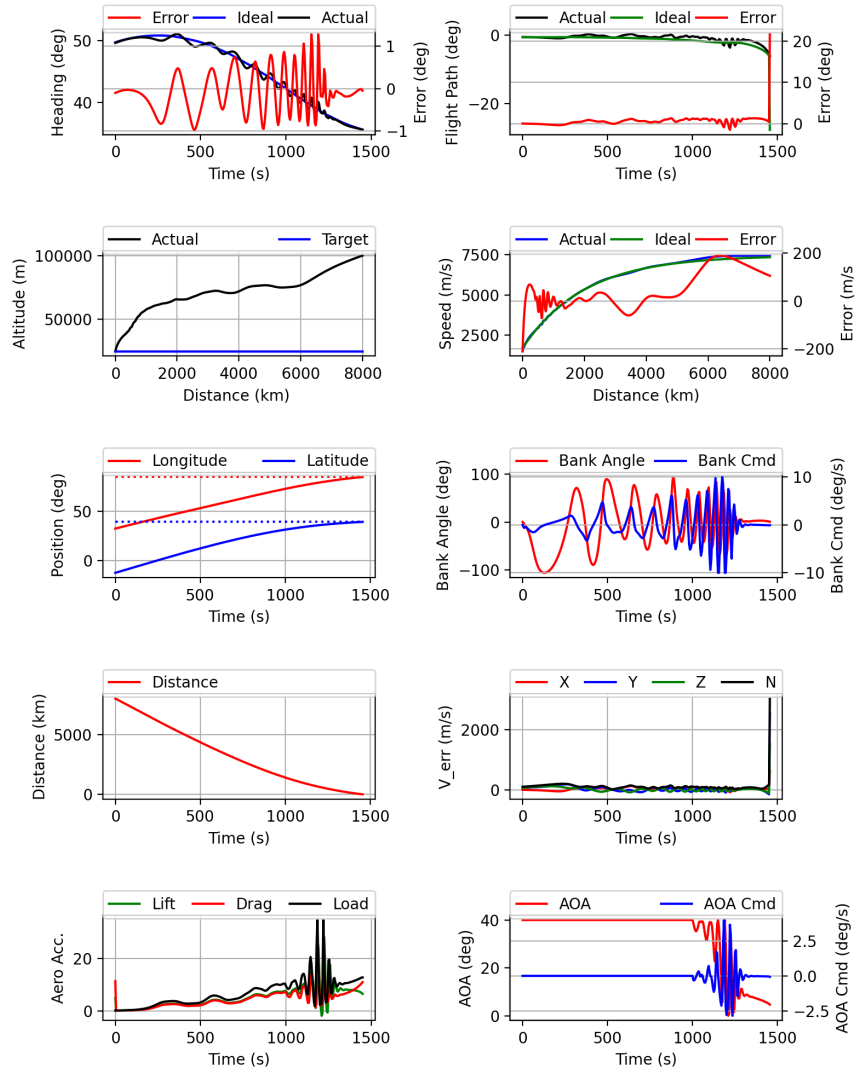

**Fig. 7   Meta-RL Guidance Sample Trajectory ("Optim" Case)**

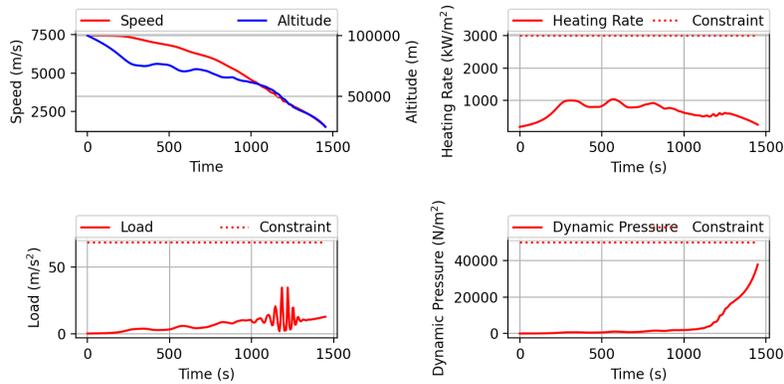

**Fig. 8   Meta-RL Guidance Sample Constraints ("Optim" Case)**

## B. LQR Benchmark

Separate trajectory generation and tracking appears to be common practice for guidance and control of hypersonic vehicles. Here we develop a linear quadratic regulator (LQR) tracker to track an optimal trajectory produced by GPOPS-II that hits a predefined final position while satisfying path constraints for heating rate, dynamic pressure, and load, as were given in Eq. (8). We used GPOPS-II to generate two reference trajectories, one for an 8000 km trajectory beginning and ending at a latitude of zero degrees, and another 8000 km trajectory beginning and ending at a latitude of 30 degrees. For both trajectories, the initial vehicle speed, range to target, and target altitude is identical to that used to optimize and test the meta-RL guidance system, as described in Table 1.

The LQR takes an observed state $\mathbf{x} = \begin{bmatrix} R & \theta & \phi & V & \gamma & \psi & \sigma & \alpha \end{bmatrix}$ and maps it to an action $\mathbf{u} = \begin{bmatrix} \dot{\sigma} & \dot{\alpha} \end{bmatrix}$. The LQR tracker requires a dedicated dynamics model with scaled variables and nominal aerodynamic parameter values. Nonetheless, once the LQR control is calculated for each guidance cycle, the simulations use the same dynamics model as for meta-RL. The LQR controller also has no knowledge of the current values of the perturbed atmospheric density and aerodynamic parameters, which would be the case in practice.

In order to perform the LQR trajectory tracking technique the equations of motion must first be linearized, and we redefine the trajectory tracking problem in terms of deviation variables of state and control given by $\delta x$ and $\delta u$ respectively. The tracking optimal control problem (TOCP) is then to find to the deviation in control $\delta u$ that minimizes the performance index $\mathcal{J}$, as given in Eqs. (21) and (26), where $A(t)$ and $B(t)$ are obtained by linearizing the nonlinear equations of motion about a reference trajectory and control.

$$\mathcal{J} = \delta \boldsymbol{x}^\mathrm{T}(t_f) \boldsymbol{P}(t_f) \delta \boldsymbol{x}(t_f) + \int_{t_0}^{t_f} \left( \delta \boldsymbol{x}^\mathrm{T}(t) \boldsymbol{Q} \delta \boldsymbol{x}(t) + \delta \boldsymbol{u}^\mathrm{T}(t) \boldsymbol{R} \delta \boldsymbol{u}(t) \right) dt \tag{21}$$

subject to

$$\delta \dot{\boldsymbol{x}} = \boldsymbol{A}(t) \delta \boldsymbol{x} + \boldsymbol{B}(t) \delta \boldsymbol{u} \tag{22}$$

From Eq. (22), $A(t)$ and $B(t)$ are approximated at discrete time points $t_k$ as constants. Thus, Eq. (22) is locally time invariant at each time constant and the linear control law is shown in Eq. (23), where $K(t_k)$ are constant feedback gains at each $t_k$. The feedback gains are found by propagating the Riccati equation backwards in time from the the final condition as shown in Eq. (24), with $K(t_k)$ is shown Eq. (25).

$$\delta \boldsymbol{u} = -\boldsymbol{K}(t_k) \delta \boldsymbol{x} \tag{23}$$

$$-\dot{\boldsymbol{P}}(t) = \boldsymbol{A}^\mathrm{T} \boldsymbol{P}(t) + \boldsymbol{P}(t) \boldsymbol{A} - \boldsymbol{P}(t) \boldsymbol{B} \boldsymbol{R}^{-1} \boldsymbol{P}(t) + \boldsymbol{Q}, \quad t \leq t_f, \quad \text{given } \boldsymbol{P}(t_f) \tag{24}$$

$$\boldsymbol{K}(t_k) = -\boldsymbol{R}^{-1} \boldsymbol{B}^\mathrm{T} \boldsymbol{P}(t_k) \tag{25}$$

Once the feedback gains at the selected discrete time points are known, they can be used to find the feedback gains throughout the trajectory via interpolation. The actual guidance commands $\boldsymbol{u}_c$ with respect to the reference values are given in Eq. (26).

$$\boldsymbol{u}_c = \boldsymbol{u}_\mathrm{ref} + \delta \boldsymbol{u}. \tag{26}$$

Any application of LQR tracking represents a trade off between hitting the terminal state, tracking the reference state throughout the time span, and mimicking the nominal guidance commands. This trade off is represented by the weighted matrices shown in Eq. (21), e.g., $\boldsymbol{P}(t_f)$, $\boldsymbol{Q}$, and $\boldsymbol{R}$. The importance of tracking each element of the reference state and mimicking the reference control are chosen by the engineer through their selection of the $Q$ and $R$ diagonal matrices. The final condition of the Riccati equation $\boldsymbol{P}(t_f)$ is a diagonal matrix that simply weights the importance of hitting the terminal state. In our study the weighting matrices were selected according to Bryson's rule [25], i.e., they were determined from maximum allowed deviations from the reference. The LQR weighted matrix values that we used are given in Table 9.

Performance is tabulated in Table 10, where statistics were generated using 5000 simulated episodes. We found that the LQR tracker's performance deteriorated significantly when the initial altitude and speed were perturbed as in the "ICV" case. Moreover, the LQR performance deteriorated significantly when modeling actuator failure. Consequently, for the results given in Table 10, the initial variation of altitude and speed were set to zero, and actuator failure was not modeled. This is indicated by the "-E" suffix (for easy) in the cases in Table 10. We found that the performance was similar between the 0 degree and 30 degree latitude trajectories, and the results shown in Table 10 are for the zero latitude optimal trajectory. Also note in all cases the actuator time constant was set to "None", which disables actuator delay, making the conditions exactly the same as that used for optimization with GPOPS. Table 11 gives statistics for the heating rate, load, and dynamic pressure constraints for the "Optim-E" case. We also ran a scenario with 1% scale factor noise, results are omitted as performance was so poor as to make the statistics meaningless. This could be important in that in the initial glide phase at high re-entry speeds, navigation might initially be inertial, and inertial measurement unit rate gyro and accelerometer bias results in increasing amounts of scale factor noise as the trajectory progresses. Note that to make a fair comparison, we did not apply the scale factor error to $R$, but instead first subtracted the earth's radius (giving us the altitude), applied the scale factor error to the altitude, and then added back the earth's radius. It is worth noting that we ran some experiments where we did not consider Coriolis force, and the results were much improved, although performance was still inferior to that of the meta-RL policy.

Table 9   LQR Matrix Values

| LQR Matrices | Values |
|---|---|
| $P(t_f)$ | $\text{diag}\left(\begin{bmatrix} 1\times 10^4 & 1\times 10^6 & 1\times 10^6 & 1\times 10^4 & 1 & 1 & 1 & 1 & 1 \end{bmatrix}\right)$ |
| $Q$ | $\text{diag}\left(\begin{bmatrix} 1\times 10^4 & 1\times 10^6 & 1\times 10^6 & 1\times 10^4 & 1 & 1 & 1 & 1 & 1 \end{bmatrix}\right)$ |
| $R$ | $\text{diag}\left(\begin{bmatrix} 1 & 1 \end{bmatrix}\right)$ |

Table 10   LQR Performance

| Case | Miss (m) | | | |Alt Err (m)| | | | $|V_{\text{err}}|$ (m/s) | | | Success | Violation |
|---|---|---|---|---|---|---|---|---|---|---|---|
| - | $\mu$ | $\sigma$ | Max | $\mu$ | $\sigma$ | Max | $\mu$ | $\sigma$ | Max | % | % |
| Ideal | 69 | 0 | 69 | 24 | 0 | 24 | 0 | 0 | 0 | 100.0 | 0.0 |
| ICV-E | 145 | 95 | 352 | 183 | 581 | 2497 | 1 | 1.2 | 22 | 89.2 | 3.5 |
| Optim-E | 1234 | 1340 | 13004 | 298 | 1522 | 8238 | 1 | 18 | 189 | 39.5 | 8.9 |
| PV=5-E | 3053 | 6090 | 232920 | 728 | 2786 | 15000 | 27 | 75 | 1192 | 19.75 | 42.1 |

Table 11   LQR Constraint Statistics tracking GPOPS ("Optim-E" case)

| Constraint | $\mu$ | $\sigma$ | Max | Limit |
|---|---|---|---|---|
| Heating Rate (KW/s) | 591 | 469 | 3307 | 3000 |
| Load (m/s$^2$) | 8.1 | 12.6 | 110.9 | 68.7 |
| Dynamic Pressure (Pa) | 3567 | 7847 | 97330 | 50000 |

## C. Discussion

A major difference between the meta-RL guidance policy and tracking an optimal trajectory is that in the first case, the trajectory itself can be adapted to the environment. Indeed, in the experiments using an LQR controller to track an optimal trajectory, we found performance deteriorated when the deployment environment differed from the optimization environment, violating constraints that were satisfied under nominal flight conditions. Interestingly, in [4], the authors track an optimal trajectory but obtain better performance under off-nominal flight conditions as compared to our LQR tracking experiments, and completely avoid constraint violations. The major difference is that in [4], a new reference trajectory is generated at each time step. Thus, the guidance system in [4] is a form of model predictive control, making it more robust to off-nominal flight conditions. Another way to view the difference between the guidance system and trajectory tracking approaches is to consider it a comparison between integrated guidance and control versus separate guidance and control systems "bolted together". In the latter case, both systems might be optimal when considered separately, but can be sub-optimal when combined into a larger system.

To better test the hypothesis that maximizing performance requires the trajectory to adapt to the environment, we optimized an adaptive tracking controller by making a simple change to the meta-RL guidance optimization environment: instead of creating the reference velocity field using the heuristics of Section IV.A, the reference velocity field is taken from the speed, flight path angle, and heading angle of a reference trajectory created using the "Ideal" scenario. By using a reference trajectory created by the meta-RL guidance system under ideal conditions, we can quantify the performance difference between an adaptive guidance system and an adaptive tracking controller tracking a static reference trajectory. We found that the adaptive guidance system (Section V.A) outperformed the adaptive trajectory tracking controller by a large margin.

Solving the 3-DOF problem posed in this paper using optimal control was computationally tractable. However, solving the problem in 6-DOF may not be possible in real time (which would be required for divert maneuvers or for re-calculating the trajectory at each navigation step) unless the problem can be convexified in a way that does not overly simplify the problem. Moreover, the tracking controller in [4] linearizes the dynamics, but when structural dynamics (i.e., body bending) are taken into account the linearized dynamics can become unstable ([26]). Thus, there are multiple advantages of the meta-RL approach as compared to tracking an optimal trajectory. First, the policy can be run forward in a few milliseconds even for complex 6-DOF problems [11, 12], and second, there is no need for constraint convexification or linearizing dynamics. Other benefits of the meta-RL approach include the ability to optimize an integrated guidance and control system, and in some cases integrate navigation as well [10]. Finally, the ability to adapt to off-nominal flight conditions has the potential to improve the performance of hypersonic vehicles.

## VI. Conclusion

We used reinforcement meta learning to optimize a guidance system suitable for the approach phase of a hypersonic vehicle. The guidance system maps navigation system outputs to commanded rates of change for the bank angle and angle of attack, and implements parallel navigation, keeping the vehicle's velocity vector aligned with the line of sight to target in curved space while satisfying path constraints. The optimized system was tested over a range of scenarios inducing off-nominal flight conditions, including perturbation of aerodynamic coefficients, sensor noise, and actuator failure scenarios. Performance was then compared to a benchmark where a linear quadratic regulator is used to track an optimal trajectory, and we found that the guidance system outperformed the benchmark by a large margin. To further investigate the performance difference between guidance system and trajectory tracking approaches, we developed a meta-RL tracking controller and used the controller to track a trajectory induced by the meta-RL guidance system. Again, we found that the guidance law outperformed the trajectory tracker. Although not demonstrated in this work, in Section IV.A we outlined a method that would allow the optimized policy to implement large lateral diverts on the order of 25% of the initial distance to target. To our knowledge this is the first published work applying reinforcement meta learning to the optimization of a guidance system suitable for hypersonic vehicles. Future work will apply meta reinforcement learning to integrated and adaptive guidance systems (six degrees of freedom) for both approach and terminal phase hypersonic vehicle guidance, and explore novel sensing and actuation methods.

## Funding Sources

This paper describes objective technical results and analysis. Any subjective views or opinions that might be expressed in the paper do not necessarily represent the views of the U.S. Department of Energy or the United States Government. This work was supported by the Sandia National Laboratories Laboratory-Directed Research and Development Program. Sandia National Laboratories is a multi-mission laboratory managed and operated by National Technology & Engineering Solutions of Sandia, LLC, a wholly owned subsidiary of Honeywell International Inc., for the U.S. Department of Energy's National Nuclear Security Administration under contract DE-NA0003525.

## References

[1] Witeof, Z., and Pasiliao, C. L., "Fluid-thermal-structural interaction effects in preliminary design of high speed vehicles," *56th AIAA/ASCE/AHS/ASC Structures, Structural Dynamics, and Materials Conference*, 2015, p. 1631. https://doi.org/10.2514/6.2015-1631.

[2] Zhang, D., Liu, L., and Wang, Y., "On-line reentry guidance algorithm with both path and no-fly zone constraints," *Acta Astronautica*, Vol. 117, 2015, pp. 243–253. https://doi.org/10.1016/j.actaastro.2015.08.006.

[3] Xu, M., Chen, K., Liu, L., and Tang, G., "Quasi-equilibrium glide adaptive guidance for hypersonic vehicles," *Science China Technological Sciences*, Vol. 55, No. 3, 2012, pp. 856–866. https://doi.org/10.1007/s11431-011-4727-z.

[4] Wang, Z., and Grant, M. J., "Autonomous entry guidance for hypersonic vehicles by convex optimization," *Journal of Spacecraft and Rockets*, Vol. 55, No. 4, 2018, pp. 993–1006. https://doi.org/doi.org/10.2514/1.a34102.

[5] Pan, L., Peng, S., Xie, Y., Liu, Y., and Wang, J., "3D guidance for hypersonic reentry gliders based on analytical prediction," *Acta Astronautica*, Vol. 167, 2020, pp. 42–51. https://doi.org/10.1016/j.actaastro.2019.07.039.

[6] Lu, P., Forbes, S., and Baldwin, M., "Gliding guidance of high L/D hypersonic vehicles," *AIAA Guidance, Navigation, and Control (GNC) Conference*, 2013, p. 4648. https://doi.org/10.2514/6.2013-4648.

[7] Yan, X., Wang, P., Xu, S., Wang, S., and Jiang, H., "Adaptive Entry Guidance for Hypersonic Gliding Vehicles Using Analytic Feedback Control," *International Journal of Aerospace Engineering*, Vol. 2020, 2020. https://doi.org/10.1155/2020/8874251.

[8] Liu, C., Dong, C., Zhou, Z., and Wang, Z., "Barrier Lyapunov function based reinforcement learning control for air-breathing hypersonic vehicle with variable geometry inlet," *Aerospace Science and Technology*, Vol. 96, 2020, p. 105537. https://doi.org/10.1016/j.ast.2019.105537.

[9] Gaudet, B., Linares, R., and Furfaro, R., "Deep reinforcement learning for six degree-of-freedom planetary landing," *Advances in Space Research*, Vol. 65, No. 7, 2020, pp. 1723–1741. https://doi.org/10.1016/j.asr.2019.12.030.

[10] Gaudet, B., Linares, R., and Furfaro, R., "Six degree-of-freedom body-fixed hovering over unmapped asteroids via LIDAR altimetry and reinforcement meta-learning," *Acta Astronautica*, 2020. https://doi.org/10.1016/j.actaastro.2020.03.026.

[11] Gaudet, B., Furfaro, R., Linares, R., and Scorsoglio, A., "Reinforcement Metalearning for Interception of Maneuvering Exoatmospheric Targets with Parasitic Attitude Loop," *Journal of Spacecraft and Rockets*, 2020, pp. 1–14. https://doi.org/10.2514/1.A34841.

[12] Gaudet, B., Linares, R., and Furfaro, R., "Terminal adaptive guidance via reinforcement meta-learning: Applications to autonomous asteroid close-proximity operations," *Acta Astronautica*, 2020. https://doi.org/10.1016/j.actaastro.2020.02.036.

[13] Mishra, N., Rohaninejad, M., Chen, X., and Abbeel, P., "A Simple Neural Attentive Meta-Learner," *International Conference on Learning Representations*, 2018.

[14] Frans, K., Ho, J., Chen, X., Abbeel, P., and Schulman, J., "META LEARNING SHARED HIERARCHIES," *International Conference on Learning Representations*, 2018.

[15] Wang, J. X., Kurth-Nelson, Z., Tirumala, D., Soyer, H., Leibo, J. Z., Munos, R., Blundell, C., Kumaran, D., and Botvinick, M., "Learning to reinforcement learn," *arXiv preprint arXiv:1611.05763*, 2016.

[16] Shneydor, N. A., "Missile guidance and pursuit: kinematics, dynamics and control," Elsevier, 1998, pp. 77,78. https://doi.org/10.1533/9781782420590.

[17] Schulman, J., Wolski, F., Dhariwal, P., Radford, A., and Klimov, O., "Proximal policy optimization algorithms," *arXiv preprint arXiv:1707.06347*, 2017.

[18] Chung, J., Gulcehre, C., Cho, K., and Bengio, Y., "Gated feedback recurrent neural networks," *International Conference on Machine Learning*, 2015, pp. 2067–2075.

[19] Joshi, G., and Chowdhary, G., "Deep Model Reference Adaptive Control," *2019 IEEE 58th Conference on Decision and Control (CDC)*, IEEE, 2019, pp. 4601–4608. https://doi.org/10.1109/cdc40024.2019.9029173.

[20] Vinh, N. X., Busemann, A., and Culp, R. D., "Hypersonic and Planetary Entry Flight Mechanics," The University of Michigan Press, 1980, pp. 20–28.

[21] Lu, P., "Entry guidance and trajectory control for reusable launch vehicle," *Journal of Guidance, Control, and Dynamics*, Vol. 20, No. 1, 1997, pp. 143–149.

[22] Finn, C., Abbeel, P., and Levine, S., "Model-Agnostic Meta-Learning for Fast Adaptation of Deep Networks," *ICML*, 2017.

[23] Schulman, J., Levine, S., Abbeel, P., Jordan, M., and Moritz, P., "Trust region policy optimization," *International Conference on Machine Learning*, 2015, pp. 1889–1897.

[24] Kullback, S., and Leibler, R. A., "On information and sufficiency," *The annals of mathematical statistics*, Vol. 22, No. 1, 1951, pp. 79–86.

[25] Dukeman, G., "Profile-following entry guidance using linear quadratic regulator theory," *AIAA guidance, navigation, and control conference and exhibit*, 2002, p. 4457.

[26] Xu, B., and Shi, Z., "An overview on flight dynamics and control approaches for hypersonic vehicles," *Science China Information Sciences*, Vol. 58, No. 7, 2015, pp. 1–19.